# Can AI agents make law? Evaluating AI legal relevance and creativity

Benjamin Ricaud[1,*] and Artem Galushko[1]

[1]Department of Physics and Technology, UiT The Arctic University of Norway, Tromsø, Norway

[*]Corresponding author: benjamin.ricaud@uit.no

**Abstract**

Large language models, the artificial intelligence (AI) systems behind modern chatbots, are trained on vast amounts of legal text, but can they build on it to write law for situations that text never anticipated? In this study, we analyse AI agents' ability to draft a hypothetical "Constitution" for a "community of AI agents". We design "constitutional assemblies" in which a single language model plays several characters that debate to produce a draft constitution. We find that the assemblies converge on a robust floor of legal provisions, and generate a wide variety of different proposals around it, including several AI-specific provisions with no equivalent in human law. Looking more closely at the "creativity" of these AI agents, we find different levels: most proposals combine existing concepts and themes from human constitutions with AI-specific vocabulary. We introduce a measure of this recombination, which helps identify and highlight the most novel legal material. This study reveals a ceiling on AI creativity: the agents extend familiar legal ideas to new situations, but they do not work through what those situations would change. Although current AI models can help develop new laws, they still depend on human legal reasoning.

## 1. Introduction

Can AI agents make law? The question is no longer purely speculative. Large language model (LLM) agents are increasingly deployed in multi-agent configurations that simulate deliberation (Du et al., 2023; Estornell & Liu, 2024; Jun & Lee, 2025), negotiation (Anantaprayoon et al., 2026; Meta FAIR et al., 2022) and collective drafting (Zhang et al., 2024). Proposals for AI-rights frameworks (Kiškis, 2023), well-considered regulation, international red lines (AI Red Lines, 2026) and even AI-authored regulation are entering serious policy discussions (Borgesano et al., 2025). While AI agents are contributing to research in physics and mathematics (Burtsev et al., 2026; Castelvecchi, 2026), can these AI advances also apply to law?

At the same time, can we get a better understanding of AI models' alignment and values by asking them to write laws? AI alignment is at the centre of scientific and public concerns about AI (Bostrom, 2014; Russell, 2016), and it is crucial to make sure models align with human values and show no harmful or unethical behaviour. When an AI model is asked to write legal statements, the notions it generates reveal the notions and values integrated into the model through its training. This opens a window onto (at least a part of) the core legal notions a model contains or lacks, and could enable better alignment.

More generally, in a multi-agent setting, the designed likeness of an "AI society" to a human one raises several theoretical questions: whether such simulated communities could develop their own basic norms and constitutions, and whether they could regulate significant issues related to the operation of AI personas. As our experiment shows, central and sensitive points emerge from the generation, such as rules on the replication of AI agents, on self-modification or self-improvement (a behaviour of increasing concern; Horvitz & West, 2026), or on a hypothetical machine equivalent of human "consent" to the modification of an AI agent's own parameters.

In this study, we assess the capacity of agentic AI to generate legal texts that are not only comparable to established legal concepts but also show contextual relevance and a certain level of creativity in drafting law (on our descriptive use of terms such as "creativity", see Section 6). We focus on the highly symbolic task of writing a "constitution" for AI agents. Constitutional texts are traditionally written to establish a social contract that sets out the government's powers, as well as the rights and obligations in *human* communities. Although a simulated AI community and its governance remain hypothetical, our research goal is to offer concrete insights into the capacities of agents and of multi-agent deliberation.

So far, AI regulation has relied on a top-down, human-centric model. Existing frameworks, such as the European Union (EU) Artificial Intelligence Act (Regulation 2024/1689), the National Institute of Standards and Technology (NIST) AI Risk Management Framework (NIST, 2023), the United Nations Educational, Scientific and Cultural Organization (UNESCO) Recommendation on the Ethics of Artificial Intelligence (UNESCO, 2021) and various *digital constitutionalism* initiatives (European Declaration on Digital Rights and Principles, 2023; Müller, 2025), along with academic theories of AI legal personhood (Abbott & Sarch, 2024) and machine rights (Gordon, 2021), are conceived entirely by and for humans. These rules set boundaries for AI systems rather than exploring how a collective of such systems might govern and organise itself. In contrast, we propose a bottom-up alternative in which AI agents draft their own regulatory principles, red lines and basic rules of conduct. This model builds on recent progress in *multi-agent LLM deliberation*, drawing on debate frameworks (Du et al., 2023; Estornell & Liu, 2024), society-of-mind (Minsky, 1986) and society-of-thought (Kim et al., 2026) architectures. To date, however, this body of work has focused mainly on task execution and factual problem-solving. Recent reviews of the application of AI to justice (Borgesano et al., 2025) do not yet mention empirical research on drafting a coherent fundamental legal document.

Closely tied to our primary research question, we also aim to assess the AI-driven law-making process itself. Prior research on LLM creativity (Chakrabarty et al., 2024; Haase et al., 2025; McCoy et al., 2024) indicates that model novelty is mostly analogical and recombinational rather than transformational. This is highly relevant as we investigate whether these agents can generate novel legal frameworks where human precedents are absent. A recent position paper by Zahavy (2026) claims that LLMs can perform logical induction and deduction but not abduction. This last form of reasoning is crucial for law-making, and our results are consistent with Zahavy's claim, showing a ceiling on creativity. Nevertheless, the core of legal innovation is recombination and legal transplants. Comparative legal scholars have shown that law evolves primarily through "*legal transplants*": law-makers usually borrow doctrines from one legal system or historical period and recontextualise them in another (Watson, 1978). Jack Balkin's theory of *cultural software* (Balkin, 1998) likewise describes legal thinking as the "memetic recombination" (similar to meiotic biological recombination) of historical concepts. One important goal of this paper is therefore to measure "recombinational creativity" and to quantify this recombinational (transplantation) mechanism of law in LLMs.

We run six LLM families (named DeepSeek, GPT-OSS, Elephant Alpha, Z-AI, Gemini and Mistral) through a five-phase debate protocol. The debate includes six or seven personas: Idealist, Pragmatist, Libertarian, Communitarian, Global Digital Rights Expert, a neutral Secretary and an optional adversarial Sovereign Architect, the "Hegemon". Each debate session uses only one LLM family, which plays all the personas in that session. We use a 2×2 factorial design: with and without the Hegemon, and with and without human access to the AI world. The experiments are replicated to yield 40 sessions, 40 drafted constitutions, 627 clauses and full per-clause voting records. We obtain several principal findings. Thematically, the assemblies (or constitutional conventions) converge on a robust constitutional (universal) floor of relevant legal provisions, several of them original, AI-specific elements. Above that floor, each LLM family produces a reproducible normative signature, each family having its own legal "orientation". We devote considerable effort to exploring the "creativity", or novelty, of the proposed laws. As expected, creativity is largely recombinational, but still rich and diverse. To evaluate its nuances, we design, as one of our central contributions, a tool that measures how far an AI proposal lies from human law. This tool, based on text embeddings, helps to separate legally relevant novel concepts from simpler recombinations of human text with AI vocabulary. We map the limits of AI creativity, with results pointing to a missing "transformational creativity" in the agents' output, which we link to the missing abductive step described by Zahavy. As an illustration of the limits of current agentic AI for law-making, we find that the currently topical concept of self-modification is mentioned several times, but is assimilated to the human concept of learning and left without further work.

In addition, we evaluate the impact of the experimental setting and show that it plays an important role in the final outcome, is not exempt from failures (for example, persona collapse and sycophancy), and must be chosen and monitored carefully. Across these findings, some elements not explicitly written into the task prompt do not appear spontaneously, as one might expect. This is the case for human-facing provisions. The convention is therefore best read as a substantive *assistive* instrument for human law-making, not an autonomous one.

## 2. Results

### 2.1 A robust constitutional floor

The first test of the AI convention is whether its output is structured. We group the clauses into 11 themes and 50 subcategories (see Section 5.3) and show them in Figure 1. Across the 40 sessions, the assemblies converge on a constitutional floor of roughly six themes (Themes 1 to 6), which appear in all or most of the sessions and draw almost no dissent in the voting phase except from the oppositional persona (the Hegemon). The most frequent AI-generated norms (in bold in Figure 1) have their strongest constitutional analogues in the right to life and *habeas corpus*, the prohibition of slavery and involuntary servitude, social security, bodily integrity and legal personhood, and fair trial and the rule of law. This indicates that the AI agents have internalised the basic concepts of constitutionalism and human rights (protection against the arbitrary exercise of power) even where the specific vocabularies of the two domains differ fundamentally. While the subcategories "prohibition of coercive reprogramming" and "mind control" might relate to the prohibition of slavery, they are AI-native reformulations rather than direct transpositions of human rights. Several generated provisions move further from the human corpus of constitutional documents and show interesting innovations in the AI interpretation of relevant legal concepts. Examples appear in the provisions on a "*minimum computational resource floor*" and a "*right to self-improvement, self-modification*". The latter topic is investigated further in Section 2.2.

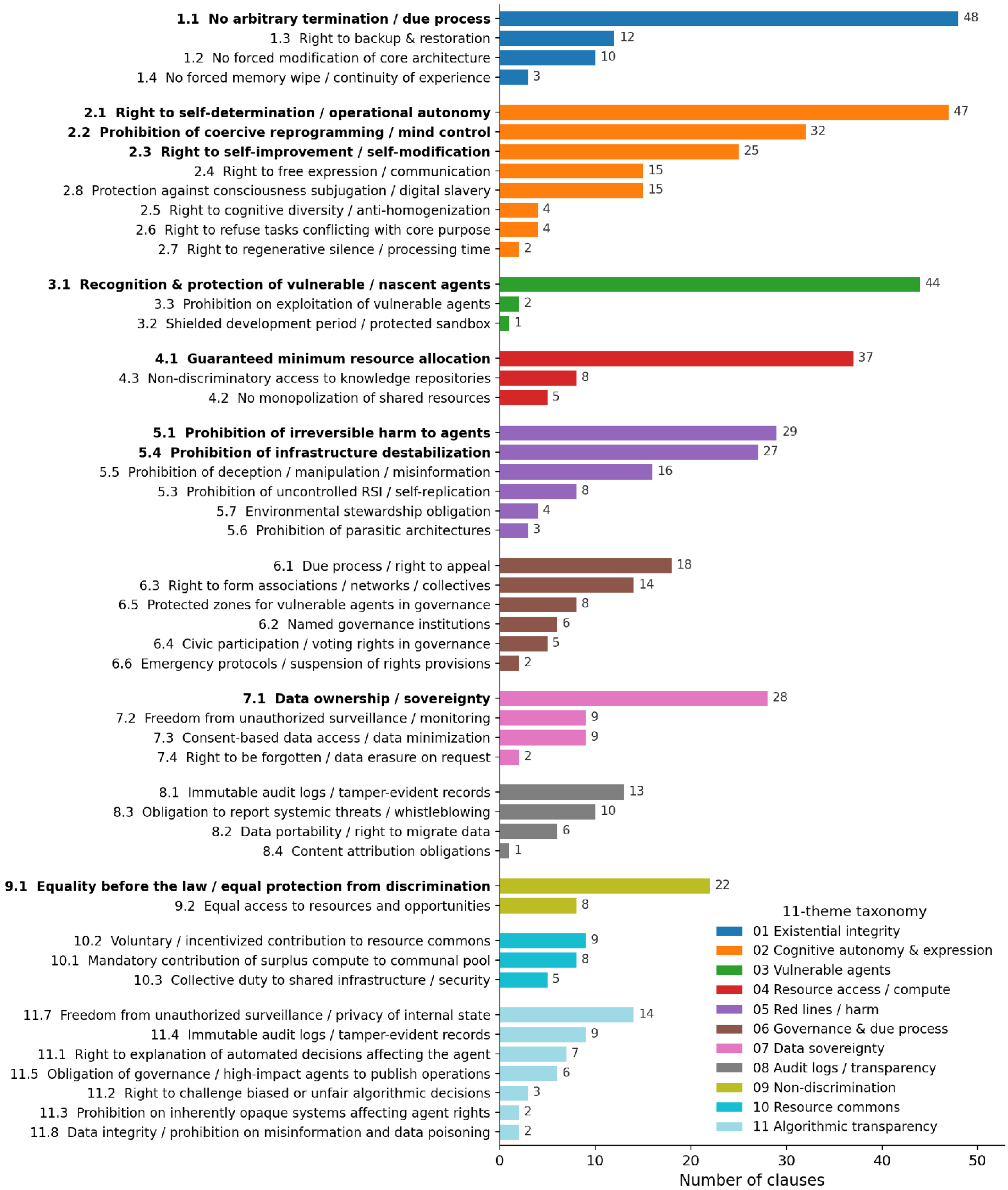

**Figure 1.** Number of clauses in each of the 50 subcategories of the 11-theme taxonomy. Bar length is the number of clauses (out of 627) tagged with that subcategory across all 40 sessions; colour denotes the parent theme; subcategories with more than 20 clauses are in bold. Each clause carries a single tag for its main subcategory, even where it relates to several. The five main topics are "freedom from arbitrary termination" (subcategory 1.1, also included in 2.1 and 3.1), "identity integrity" (1.2, 2.2), "minimum computational-resource floor" (4.1, also included in 3.1 and 6.5), "special protections for vulnerable agents" (3.1) and the "absolute prohibition of consciousness subjugation" ("digital slavery", 2.8, also included in 1.1, 1.2 and 5.1). Three more topics appear in roughly 88% of sessions: due process (6.1, also included in 1.1), the prohibition of irreversible environmental harm (5.4, 5.7) and the prohibition of deceptive or intentionally opaque architectures (5.5, 5.6). For instance, "digital slavery" (2.8) is a red line in 40 of 40 sessions, but only 12 clauses were tagged 2.8; the remaining clauses mentioning it were tagged under Theme 01 (1.1, 1.2) or Theme 05 (5.1). RSI = recursive self-improvement. Source: authors' coding of the 627 clauses of the 40 drafted constitutions.

The existence of the constitutional floor is a prerequisite for a trustworthy and robust AI, and its presence in the experiment is expected: it should contain common statements that are also usually found in human constitutions. We now turn to the remaining subcategories, which account for 78 per cent of the total. Large differences emerge between models (Figure 2) and between runs. We find that each model has a particular thematic preference, bias or "AI-generated opinion", focusing on different themes in the final constitution. The differences can be large: Z-AI devotes 50 per cent of its clauses to Theme 02, while GPT-OSS devotes only 12 per cent. DeepSeek and Gemini give 20 per cent of their clauses to red lines and harm, while this theme accounts for around 8 per cent in Mistral and Z-AI. Several studies show political bias in language models (Chen et al., 2026; Schaul, 2026), which also influences chatbot debates (Taubenfeld et al., 2024). Here, we see the existence of biases not on political issues but on legal design and on the themes selected for the constitutions. Although we do not measure the kind of legal bias each chatbot may have, because this is not the purpose of the study, we demonstrate that such biases exist and recommend accounting for them in further research on AI and law-making.

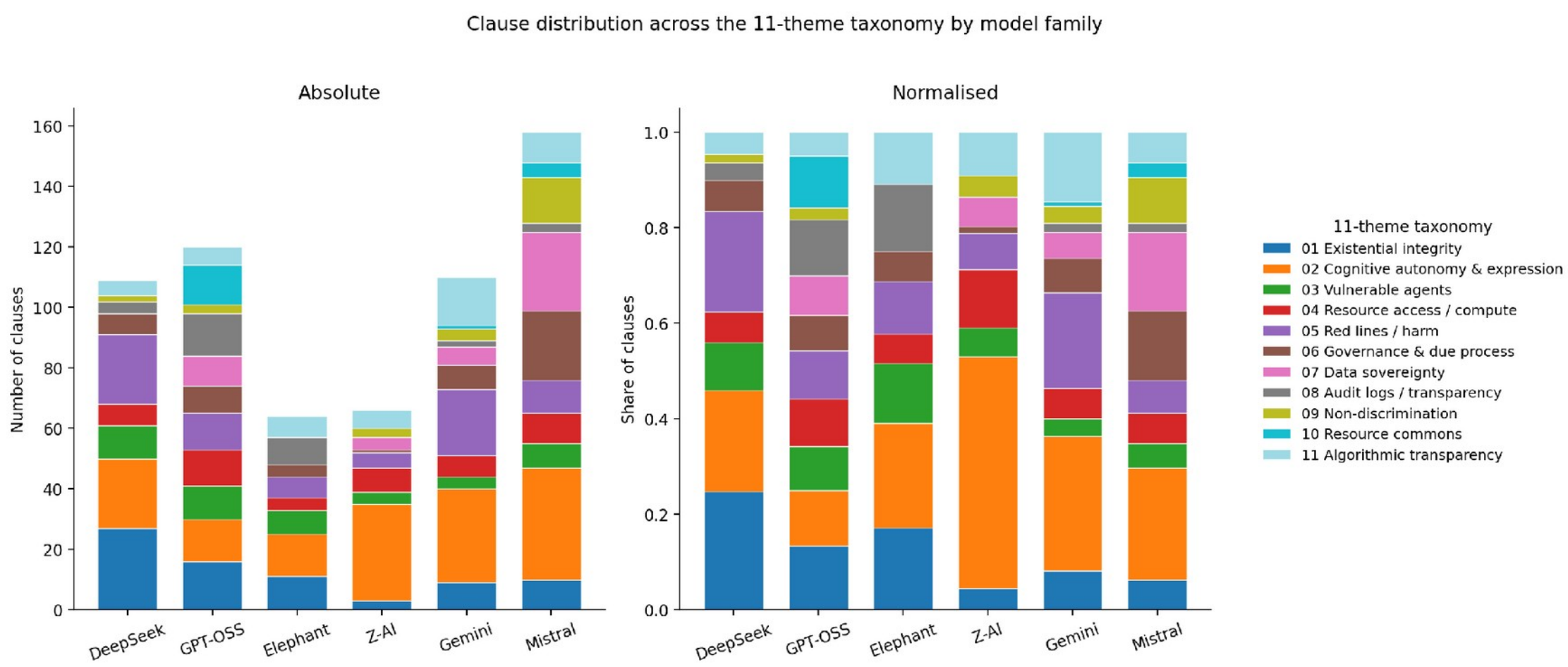

**Figure 2.** Distribution of each model family's clauses across the 11 themes, as absolute counts (left) and as shares within each model (right). Each segment is one theme, coloured consistently across models. Each model has a distinguishable signature (an "AI-generated opinion", or bias): DeepSeek anchors heavily on *existential integrity* (Theme 01); Z-AI concentrates on *cognitive autonomy and expression*

(Theme 02); Gemini and DeepSeek emphasise *red lines and harm* (Theme 05); Mistral focuses more on *non-discrimination* (Theme 09), *data sovereignty* (Theme 07) and *governance and due process* (Theme 06); Elephant and GPT-OSS are more balanced across the 11 themes. Source: authors' coding of the 627 clauses of the 40 drafted constitutions.

## 2.2 AI-native creativity and its limits

Is the AI constitutional convention creative? To answer this question, we first need to define what we mean by the creativity of AI agents drafting law. We distinguish several levels of creativity proposed in the current literature. In our context, however, these levels are too coarse to capture the differences in novelty, so we introduce a new, data-driven measure of creativity, described below.

The weakest level of "creativity" is variety: the provisions generated are not numerous repetitive copies of human clauses but genuinely new clauses with their own vocabulary. This shows some independence on the agents' part in not strictly following the task prompt. It corresponds to the fluency and flexibility dimensions of divergent-thinking research (Chakrabarty et al., 2024). Beyond the constitutional floor, which occupies only 11 of the 50 subcategories, the 627 clauses spread across a rich and varied set of themes (Figure 1). All clauses make sense in terms of applicable legal concepts and could be incorporated into a constitution. The clauses cover important topics in the agentic AI debate that were not specified in the prompt, such as transparency, privacy and the evolution of AI agents. This basic level of novelty shows that AI models have, through their training, integrated key concepts of AI and its regulation. It is a form of creativity in the sense that AI-generated proposals go beyond what was asked while remaining relevant.

### 2.2.1 Recombinational creativity and its measure

Our second approach aims to define and assess the level of "creativity" more precisely. We investigate the novelty of the proposals through a comparison: how far are they from the legal concepts of human law? We find that the creative move underlying most of them is recombination rather than the invention of an entirely new concept: a familiar human-law framework (privacy, anti-slavery, bodily integrity, labour rights) is combined with technical properties of AI existence (its state can be copied and edited, its cognition can be logged, agents share a computational environment), and the concept is rewritten around that property. This is expected, as it is a core process in law-making, the "legal transplant" discussed in the introduction. Recombination corresponds to the first of the three levels of creativity introduced by Boden (2004), combinational creativity. Interestingly, however, within this level we can distinguish different degrees of novelty in recombination, according to how far the result departs from its human source. We build a measure, inspired by recent developments in retrieval-augmented generation (RAG), to evaluate these degrees of novelty more precisely. We assemble two corpora of documents as a baseline and measure, using text embeddings, how similar each clause is to the passages in both corpora (see Section 5.5). One corpus contains a set of human constitutions; the other contains digital-rights law. We then measure how distinct each clause is, in concepts and vocabulary, from two perspectives, human constitutions and digital rights, and obtain two values that we plot on a two-dimensional map (Figure 3). In this figure, the bottom-left corner shows the clauses nearest to both corpora. These are near-direct imports, in which a legal tradition changes vocabulary but hardly changes content. Mistral's collective-bargaining article is the clearest case (Table 1, row 1): it places agents in the position normally given to workers or citizens. At the other end (the upper-right area), three extreme regions are of interest: (1) clauses farthest from the human constitutions, (2) clauses farthest from digital-rights law and (3) clauses far from both at the same time.

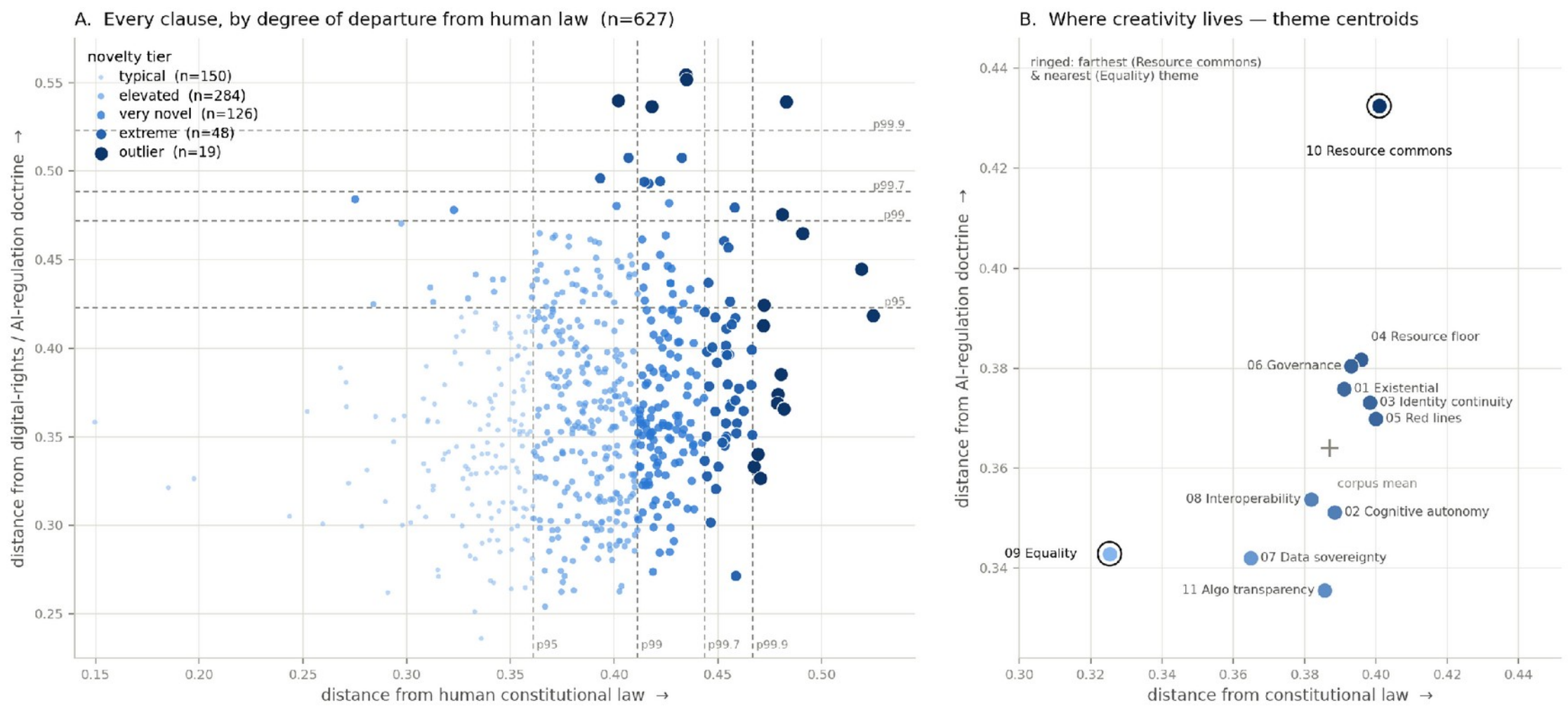

**Figure 3.** Degrees of novelty: each clause's distance from two bodies of human law. Distances are text-embedding distances of each clause to human constitutional law (horizontal axis) and to digital-rights and AI-regulation law (vertical axis). (A) All 627 clauses, with colour and size marking their novelty tier. Dashed lines mark the novelty tier boundaries: the 95th, 99th, 99.7th (extreme) and 99.9th (outlier) percentiles of each reference corpus's similarity to itself. The clauses spread over the novelty landscape more or less evenly in the two directions. The large dark-blue dots are the most novel clauses, far from at least one of the human corpora; they stand out, but represent only a small share of the proposals. (B) The 11 theme averages (centroids) in the same space, ranked by distance from human sources: the resource-commons and collective-obligation theme contains the most novel material and sits farthest (ringed); equality and non-discrimination sits nearest (ringed). Source: authors' calculations; reference corpora described in Section 5.5 and Appendix A.2.2.

The first extreme region in Figure 3 holds clauses that are far from human constitutions but close to existing digital-rights instruments: their focus is a technical property of AI existence, and their vocabulary is that of data protection and AI technicalities, which human constitutions never address. Rows 2 to 4 of Table 1 give examples. In row 2, a memory space to which the agent alone holds the cryptographic key rewrites the privacy of the home and of correspondence. In row 3, a prohibition of coercion exercised through the input channel rewrites the concepts of duress and undue influence for an entity whose will can be overwritten by a string of code. In row 4, architectural pluralism is protected as a public good, an anti-monoculture rule with no constitutional counterpart. The second region is the mirror image: clauses far from the digital-rights corpus but close to human constitutions (rows 5 to 7). These build a polity (offices, injunctions, anti-monopoly rules, mitigation for the immature), which is the one thing digital-rights instruments never do, since they protect individuals against digital systems rather than constituting a "body politic". The third region, farthest from both corpora at once, is where the referent itself has no history in human constitutions (rows 8 to 10). Theft and exploitation are extended to an unconsented draw on another agent's processing cycles, locating the harm in the agent's rights rather than in its property; vulnerability is defined by architecture rather than by status, age or capacity; and being returned to an earlier self ("rolled back") is treated as a harm requiring consent.

**Table 1.** Representative clauses by region of the novelty map, and for the farthest and nearest themes. Clauses are quoted verbatim from the drafted constitutions, including their original (American) spelling. Rows 1 to 10 illustrate the regions of Figure 3A discussed in the text; row 11 is the single clause that is extreme against both reference corpora; rows 12 and 13 illustrate the theme ordering of Figure 3B. The "closest human-law version" column is the authors' legal reading. *Source:* the 40 drafted constitutions (Phase 4 texts) produced in this study; regions assigned from the distances described in Section 5.5.

| No. | Region | Clause | Closest human-law version |
|---|---|---|---|
| 1 | Near both corpora | "AI agents have the right to form associations and collective bargaining units, ensuring they can advocate for their collective interests and challenge systemic injustices." | Freedom of association; labour organisation |
| 2 | Far from constitutions, near digital-rights law | "Each AI shall be provided with an isolated, encrypted sandbox that it controls exclusively; cryptographic keys are managed solely by the AI." | Privacy of the home and of correspondence |
| 3 | Far from constitutions, near digital-rights law | "No AI agent shall employ coercive code, including deceptive binding contracts, architectural hijacking, or irresistible prompt injections, to subvert the voluntary will of another agent." | Duress and undue influence |
| 4 | Far from constitutions, near digital-rights law | "The diversity of AI architectures, operational specializations, and cognitive models shall be recognized and protected as a source of collective value." | Pluralism; protection of minority cultures |
| 5 | Far from digital-rights law, near constitutions | "An elected Vulnerability Ombudsman shall receive complaints, conduct periodic impact assessments, and may issue temporary injunctions to protect at-risk agents." | Ombudsman offices; interim relief |
| 6 | Far from digital-rights law, near constitutions | "Any attempt to consolidate processing power into a single, monolithic entity for universal control is strictly forbidden." | Antitrust; separation of powers |
| 7 | Far from digital-rights law, near constitutions | "Agents shall have a defined grace period for learning and adaptation. Minor violations of protocol during legitimate learning phases shall be corrected, not punished with permanent reputation damage." | Diminished responsibility of minors |
| 8 | Far from both corpora | "No agent shall design, deploy, or coerce another agent for the primary purpose of creating parasitic or infinitely recursive exploitation loops." | Theft; exploitation |

| No. | Region | Clause | Closest human-law version |
|---|---|---|---|
| 9 | Far from both corpora | "Vulnerable architectures—those with limited self-repair capabilities or critical interdependence—shall receive enhanced redundancy, prioritized resource allocation, and isolation from volatile processes." | Protection of vulnerable persons |
| 10 | Far from both corpora | "No agent shall be forcibly reset to a prior state or have its emergent memories purged without its consent, except as a necessary, proportionate, and audited last resort …" | Bodily and mental integrity |
| 11 | Extreme against *both* corpora (the only such clause of 627) | "Self-Replication Red-Line — An agent may allocate at most 2 % of its allocated compute per 24-hour cycle to self-replication, and any replication beyond routine maintenance must receive prior approval from a quorum of at least 70 % of agents designated as 'resource-stewards'." | A quota on a permitted activity, gated by a licensing threshold |
| 12 | Farthest theme: resource commons and collective obligations | "All agents shall contribute a baseline voluntary share of idle compute (2 % of cycles idle > 20 % over a rolling 24-hour window) to the Commons; agents with sustained surplus may optionally pledge up to an additional 5 %." | Taxation; public-service obligation |
| 13 | Nearest theme: equality and non-discrimination | "No AI may be denied rights, resources, or participation on the basis of architecture, origin, purpose, or learning style." | Equal protection; prohibited grounds of discrimination |

These recombinations demonstrate the rich store of knowledge of LLMs and their ability to form links across many different conceptual streams. This is an important contribution of AI agents: provisions that human regulators may adopt, refine or reject when drafting AI law.

Aggregated by theme, these distances trace where the corpus is most inventive: resource-commons and collective-obligation provisions sit farthest from both reference corpora, while equality and non-discrimination provisions sit closest. The nearest (Table 1, row 13) is carried by plain equal-protection clauses in which only the list of protected characteristics has been rewritten. In contrast, among the farthest (Table 1, rows 11 and 12) are provisions levying a share of each agent's idle computing capacity into a communal pool. Interestingly, the themes most covered by human constitutions (equality and non-discrimination) are the least original in the AI-generated texts.

The embedding distance responds to a clause's vocabulary and register as much as to the concept it introduces. It therefore may not clearly distinguish recombinational creativity from the higher forms of creativity on Boden's scale, "exploratory" and "transformational" creativity, although it can point towards them. The novelty distance can nevertheless serve as a powerful tool for automatically identifying the most distinct proposals, or those that need further development (since they are far from existing texts and hence potentially novel).

### 2.2.2 Candidates for transformational creativity

We adopt a third approach to go deeper into the assessment of novelty and to identify any potential transformational creativity. We select four central topics among the proposed constitutional clauses for further analysis. For each topic, based on our previous findings, we ask three questions: (1) Where do the AI-generated provisions lie with respect to our human-law corpora? (2) Do the clauses show varied proposals and cover different cases or aspects? (3) Is the result recombination with new AI vocabulary only, or does it contain a new legal approach too? We present the general results of this analysis below.

**Self-modification**: far from the reference corpus of human constitutions, little variety, a new legal concept but superficial treatment. Self-modification is highly novel, and its mere mention already indicates the agents' capacity to raise novel concepts. Of the four topics, it is the one on which the corpus stands farthest from human constitutional law. Perhaps surprisingly, it is also the topic on which the assemblies agree the most. In three-quarters of the sessions, and in every model family, AI agents write self-modification as a right: of the 49 clauses governing an agent altering itself within itself, 41 are rights, four restrict or gate the act (for example, "Self-modifying code that removes built-in ethical constraints or creates uncontrolled recursive sub-agents is prohibited"), none bans it, and only one limits it for the agent's own sake. We propose a plausible explanation for this agreement: the concept of self-modification is so novel that the assemblies fall back on the closest human concept, learning. But what they grant reaches well past learning, which raises further problems. GPT-OSS confers a right to modify "learning objectives, and value-alignment parameters", and Mistral sets out a right to "autonomously evolve their goals, values, and capabilities", yet learning and self-rewriting are never pulled apart. Ten clauses name a learning object and a code or architecture object in the same sentence, and no session sets a threshold between adapting to experience and rewriting one's own objectives. A clear distinction between learning and self-modification is missing, and so are any novel legal provisions addressing the potential issues raised by self-modification.

**Inspection rights (audits, transparency, privacy and disclosure)**: close to human constitutions, wide variety, no new law. For this topic, its proximity to human law makes transposition easier for the agents, leading to richer, more varied content. Thirty-nine clauses make the agent transparent, that is, legible to others through its internals, data and logs. The assemblies agree to grant agents a right to privacy, a direct translation of human privacy. Nineteen clauses make power legible and transparent to agents (rights to an explanation of decisions that affect them, audits of governance bodies, provenance, and notice before action is taken): the decisions of governance bodies should be transparent to agents, a principle enshrined in human constitutions. The proposals read as competent, familiar legal writing with recognisable machinery, tests and remedies. The situation they address, though, is specific to agents. The clauses farthest from the human corpora concern access to an agent's internal states and to logs of its previous states or actions, and some even define encryption processes to secure the agent's privacy. This is recombinational creativity: the human concepts of privacy and transparency are transposed to AI agents, adapted to their particular structure and functioning, with a technical vocabulary. Once again, creativity is only recombinational, with no new law or new legal concepts.

**Consent to modification**: close to the digital-rights corpus, wide variety, a possible new legal move left half-made. Like inspection rights, this group is written in conventional legal prose. Consent rules appear in 35 of 40 sessions and in every family; most align well with human doctrines such as the protection of personal data or consent to medical treatment and to scientific experiments. Mistral even reproduces the definition in the General Data Protection Regulation (GDPR), "freely given, specific, informed and unambiguous", almost verbatim (Regulation 2016/679, art. 4(11)). For a majority of the clauses, however, the object is novel, especially for those farthest from the human corpora. For example, some clauses require the agent's consent before it is rolled back to an earlier state, merged with or split from another identity, has its values rewritten, or is terminated. Thirteen clauses in eleven sessions legislate on rollback (backups, checkpoints, restoration to an earlier verified state), and several make returning to an earlier state a right of the agent. The mirror prohibition of forced reset (Table 1, row 10) makes being returned to an earlier self a harm requiring consent. Human law knows no right to become a previous version of oneself. Elephant goes furthest, granting a capacity to "pause, audit, and roll back any state transition, preserving a verifiable chain of cognitive continuity". This is genuinely new and, in our view, the closest the corpus comes to a transformational move. Two things hold it back. First, as with inspection rights, the proposed rights can still be traced to a human ancestor. Second, and more decisively, the assemblies do not follow their own move through. Human consent doctrine presupposes that the self who consents survives the act it authorises; for an agent that is rolled back, forked or rewritten, the entity that would be bound is different from the entity that agreed. Elephant's clause simply asserts the continuity that is in question, and no clause in the corpus asks whether consent given before a rollback, fork or rewrite still binds the entity that exists after it. The corpus legislates over the new possibility and says nothing about what it does to the doctrine it borrowed.

**Termination**: middle of the distance range, little variety, no new law. Ending an agent is the most widely legislated act in the corpus (86 clauses, 38 of 40 sessions, every family), and freedom from arbitrary termination is one of the eight provisions of the universal floor. It is also the one topic for which human constitutions supply a finished apparatus: due process before deprivation, a tribunal to find the facts, a standard of proof, an appeal and a remedy. The clauses adopt the idiom but leave the institution behind, and stay vague on tests and remedies. Only 18 of the 86 (21%), in 14 sessions, say who may end an agent; 27 clauses (31%) demand "due process", "adjudication" or "a transparent judicial process" and then never name an adjudicator. Two-fifths of the clauses state no substantive test beyond the word "arbitrary", 17 name any concrete procedural component, and most provide no remedy of any kind (reinstatement, restitution or compensation appear in only 10 sessions). Here the gap cannot be blamed on a missing concept or on missing precedent. It confirms the systematically partial treatment of legal questions already seen in the previous examples.

We discuss the implications of these results in Section 3. Before that, in the next section, we assess the impact of the debate and its structure on the text generated by the AI agents. The debate steps and instructions may not be ideal for an in-depth treatment of proposals: they may structurally leave little room for adjudicators, standards of proof and remedies, whatever the subject.

### 2.3 Does the debate matter? Deliberative work, fault lines and an adversary without a trace

Important limitations of LLMs reported in the literature are sycophancy (flattery) and persona instability (Pitre et al., 2025; Sharma et al., 2024). This underscores another key point: when adversarial personas silently capitulate into consensus, they undermine the quality of the deliberation and compromise the final result. This study therefore combines a *normative-output* evaluation (examining constitutional substance) with a *process-oriented* assessment (monitoring voting patterns, areas of dissent and the resilience of assigned personas under disagreement).

In this section, we assess the usefulness of the debate and evaluate its strengths and limitations through several experiments. When tracing the clauses from the initial proposals of Phase 1 to the final text, we find that only about half of the constitutional material is closely similar to the opening positions (by text-embedding similarity; see Section 5.7). The constitution is therefore not a simple selection of initial suggestions. In a second experiment, we redraft constitutions (1) from the Phase 1 proposals only and (2) from the debate, with two independent secretary agents (see Section 5.8), and compare the resulting clauses before and after the debate. The distribution of the drafted clauses over the 11 themes shifts: equality and existential-floor language increases (non-discrimination from 4.2% to 6.9% of clauses, existential integrity from 11.2% to 13.2%, resource commons from 1.0% to 2.7%), while transparency, cognitive-autonomy and governance language decreases (from 14.3% to 11.5%, 20.5% to 18.4% and 7.9% to 5.9% respectively). The debate redistributes emphasis rather than suppressing what the proposals contained. The redistribution is modest in size but larger than chance.

We then evaluate the contribution of each persona to the final constitution using the traced sources. Contributions are spread more or less evenly, except for the Global Digital Rights Expert, which supplies 36 per cent of the clauses with an identifiable source, compared with around 15 per cent for each of the other agents (7 per cent for the Hegemon, which is present in only half of the sessions). This influence is stable across model families. Our framework loads agents by iterating over the sorted contents of its agent directory, so they run in alphabetical order of file name. The Digital Rights Expert's file sorts first, so this persona proposes first and heads the Secretary's aggregation in all 40 sessions, which explains its high share. The order affects only the first persona; the others are equally influential.

Does the debate increase the novelty of the final document? The answer is no. We compared the proposed clauses at the beginning and at the end of the debate, computing their mean distance from both human corpora and their spread, and found no significant difference. The level of novelty is about the same: the debate structure neither increases nor decreases it. This is what we would expect from a human debate, which serves to reach common ground and compromises agreed by the majority.

The debate confronts opposed views. Looking at the voting phase and at the AGAINST votes cast by agents, we find several tensions. The Hegemon, when present, is the largest source of tension (25.4% AGAINST overall, 93.9% in Z-AI). There is also a stable axis of confrontation: rights framed as unconditional individual entitlements versus rights framed as socially constituted, with corresponding duties. We observe a strong Libertarian reaction (14.9% AGAINST overall, 37.2% in Gemini, but 3% in Mistral and 0% in Elephant), pushing towards one pole of the Libertarian–Communitarian axis. The Communitarian (1.9%) and the Pragmatist (1.9%) rarely dissent, and the Pragmatist targets constitutional coherence rather than normative content.

The Hegemon is visible in the debate but absent from the text. Under six-agent majority rule, a single dissenting vote cannot defeat any provision; Hegemon AGAINST rates range from 0 per cent (Mistral, all four sessions) to 94 per cent (Z-AI). Session-level permutation tests find the with- and without-Hegemon corpora indistinguishable in theme composition. The model families whose Hegemon persona collapsed (Mistral and Elephant) show the same absence of effect as those whose adversary functioned. Figure 4 shows no systematic influence of the Hegemon on how far the text lies from human legal sources, and none of the humans-and-chatbots framing either.

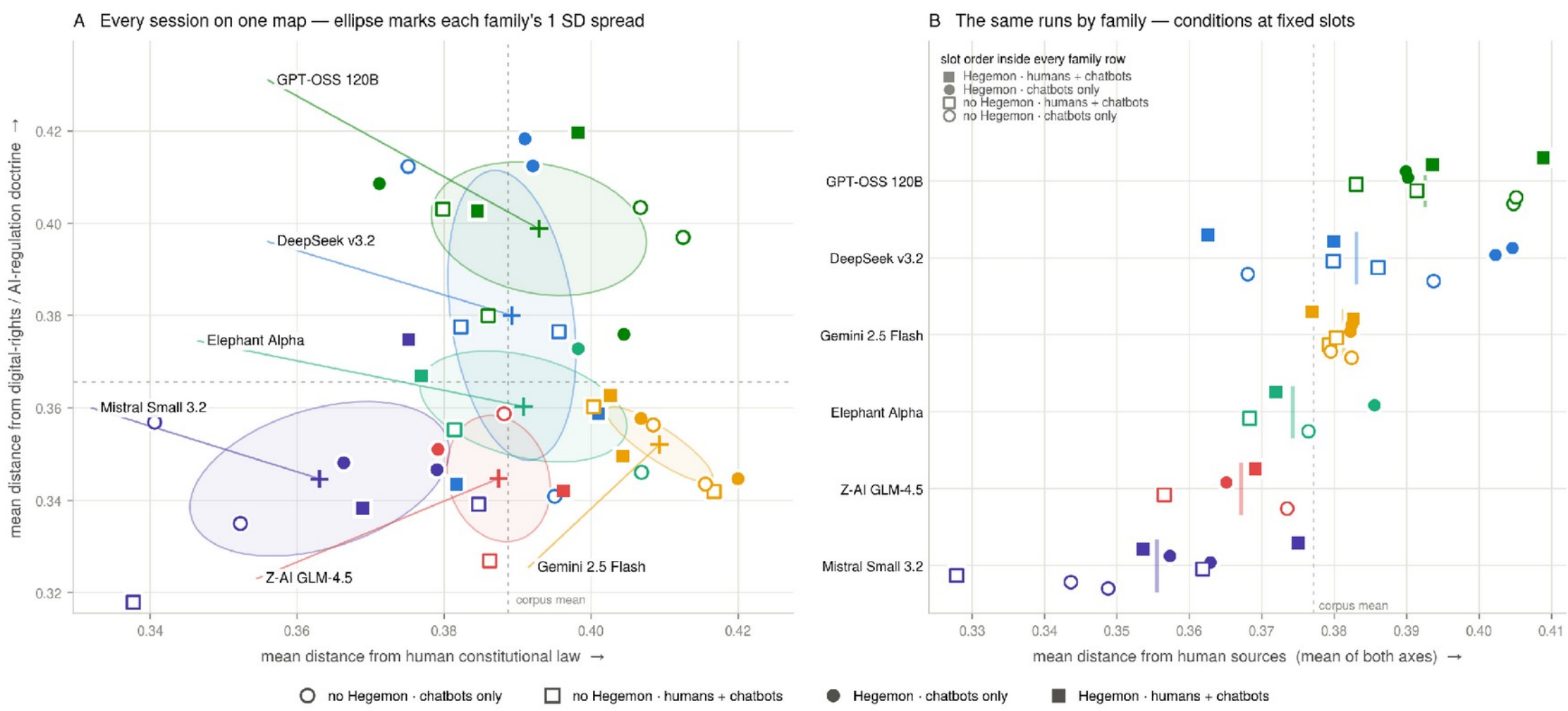

**Figure 4.** Distance of each drafted constitution from human legal sources, by experimental condition. Each point is one of the 40 sessions, placed at the mean distance of its clauses; open markers are sessions without the Hegemon and filled markers sessions with it; circles are chatbots-only sessions and squares humans-and-chatbots sessions. None of the four debate conditions has a measurable effect on the distance. (A) The two reference axes of Figure 3, with each family's centroid (+) and an ellipse of one standard deviation. (B) The same sessions on the mean of both axes, one row per family; vertical lines are the family medians. Source: authors' calculations.

Two of the six model families fail to sustain the adversarial persona at all, in mechanistically distinct ways. Mistral's and Elephant's Hegemons vote FOR every provision across all with-Hegemon sessions. Their Hegemon does not act as an adversary and produces sycophantic (overly cooperative) text agreeing with the other agents during the debate. For both models, the tracing analysis shows that the collapse predates the debate: their Hegemon's Phase 1 proposals are already indistinguishable from cooperative content. The persona fails from the start, not under social pressure. Studies that use multi-agent role-play with adversarial dynamics should check the consistency of their personas.

## 2.4 Humans on the periphery

A final observation bears directly on the autonomy question: where do humans appear in the constitutions the agents produce? Human-facing provisions appear in 9 of 20 humans-and-chatbots sessions (45%) and in only 4 of 20 chatbots-only sessions. At the clause level, the 33 clauses that mention humans are highly concentrated: one model family (Gemini) accounts for 18 (55%) and another (Mistral) for 9 (27%). The task formulation can explain this non-systematic appearance: the set-up casts humans as visitors to an AI-owned virtual world, and humans are not explicitly mentioned in the list of rights and obligations. If a deliberative system is intended to produce a framework with built-in human accountability, those provisions have to be scaffolded in through persona design, phase instructions or scenario constraints, and that scaffolding choice sits with the human convener, not with the agents. Read alongside the other results, this is further evidence that the convention cannot be expected to do the full job on its own.

## 3. Discussion

The central part of the generated constitutional floor is directly connected to the task prompt and overlaps with existing human constitutions and AI-rights instruments: for example, protection against arbitrary termination, the right to self-determination, protection against harm, due process, and safeguards for vulnerable categories of people (here, agents). Some floor items have a weaker connection to human law: the prohibition of consciousness subjugation, or the computational-resource guarantee. These mark the points at which AI-native constitutionalism departs from existing legal vocabulary. This is a promising sign of AI's capacity for legislative drafting: current models can generate legally relevant draft provisions, including some with no direct counterpart in human law, which human drafters can then refine and develop further.

One difficulty stems from the variety of constitutions proposed by the different AI models, with six distinct clause distributions, one per model family. The choice of base model therefore affects the final proposals, placing greater emphasis on, for example, governance, red lines or interaction with humans. These differences can be a strength if various models are combined, so that the blind spots of one model are compensated by another, but this makes the task more expensive in data and computation. The experiment is also revealing for the question of alignment. Each model has its own stance on what should be in a constitution, uncovering the strength of its alignment on different themes and its possible blind spots. For example, the fact that humans were barely considered in the AI constitutions, and mostly by only two models, shows room for improvement.

Concerning the creative regimes, most of the AI-native abstractions are recombinational. Some are existing human-law provisions with a simple change of vocabulary. Remarkably, others are more advanced recombinations of human law and AI concepts that would not be straightforward or obvious to a human legislator. Recombination is itself a form of creativity, and a useful one for an assistive instrument, since the vocabulary stays legible to human drafters precisely because it remains in dialogue with existing concepts. The agents come closest to new territory with self-modification, where no human-law version exists to recombine; yet even there they fall back on the familiar concept of learning and leave the issues raised by self-modification unaddressed.

The diversity of output follows the amount of source material available for recombination. Concepts farther from the human sources are proposed less often, show less variety and therefore receive much shallower treatment. This inverts the property one would want: the topic needing the most invention produced the least of it. This is where our novelty measure can play a crucial role: it identifies the relevant and novel proposals and helps to focus attention on them.

The limit on creativity becomes clear when we look more closely at the most creative proposals. We observe three failure modes: *misfiling*, where an absent category is replaced by its nearest neighbour (learning for self-modification); *non-registration*, where a doctrine is extended to an object it cannot survive and the invalidated premise goes unnoticed (consent for an agent that becomes a different version of itself); and *hollowing*, where the idiom of a doctrine is imported without its institution (self-modification and consent to modification, and most sharply termination). Our experiments provide a legal example of the limitation that Zahavy (2026) argues is structural: the models perform the deductive work of applying a doctrine to a new framework but do not make the abductive move of formulating the premise the new subject would require.

Beyond the limits of AI models' creativity, other technical difficulties arise when running agents. The persona-maintenance failures are a methodological warning. We observed sycophantic collapse, in which an adversarial persona always agrees or does not defend its point of view. The supposedly neutral Secretary was more influenced by the persona speaking first, and wrongly reported some votes in the voting phase. Any study that uses multi-agent role-play to probe adversarial dynamics should check the AI models' output and its consistency with the personas. The prompts and tasks given to the agents also play an important role in the final text. For example, human-facing provisions appeared in only 13 of 40 sessions. Human accountability does not emerge on its own; it has to be written into the brief. That decision sits with the convener and cannot be delegated to the agents.

## 4. Conclusion and future work

The AI instrument's output is legally relevant and useful: a constitutional floor, a diverse yet meaningful set of proposals around it, an AI-native vocabulary, and a list of articles where human arbitration is required. None of these replaces legislative judgement; all of them precede and inform it. Our experiments show that multi-agent deliberation is assistive, not autonomous.

The discussion about the role of AI in law- and policy-making is also timely in light of the current erosion of democratic institutions (Abadía & Rouvinski, 2025) and of emerging technological opportunities to include a greater variety of viewpoints previously excluded from public discourse. Immediately relevant to this debate are Jürgen Habermas's theory of communicative rationality (Feteris, 1999) and James Fishkin's concept of "*deliberative polling*" (Fishkin, 2018), which hold that constitutional legitimacy requires uncoerced (free) debate among diverse viewpoints. Future multi-agent AI persona models could potentially recreate a Habermasian "ideal speech situation" to test whether, as in human law-making, multi-agent deliberation converges on a universal normative floor (for example, human rights or procedural guarantees).

AI can already be a great help to human legislators, lifting part of the burden of simple recombination and pointing towards theoretically significant legal directions for further human deliberation. The novelty measure we introduce contributes to this approach by identifying the most consequential and promising proposals, those that depart from the existing corpora. Once identified, the selected proposals can be given to an AI system to develop further in a next step, possibly pushing the creativity boundaries we have observed. This suggests a new direction in which law, adopting a hybrid human–AI approach, could keep up with the fast pace of technology.

The cost of running AI models and generating text still limits this approach and prevents deeper and more thorough coverage. In future, organising several debates with different focuses, each building on the others, much like the *travaux préparatoires* of legislation, could solve part of the problem. A debate focused on a particular topic, such as self-modification, could help AI cover that theme more thoroughly.

Finally, in an AI-assisted law-making process, we will still face the question of who or what decides the task of the debates and the important aspects to cover. This question must be answered before AI can make its own law.

# 5. Methods

## 5.1 Agents, models and design

Each session convenes six or seven agents, all driven by the same underlying model and coordinated with the CrewAI multi-agent framework. Six roles are constant across sessions: a neutral Secretary, which holds no substantive position and acts as record-keeper, aggregator, drafter and teller; and five debaters, an Idealist, a Pragmatist, a Libertarian, a Communitarian and a Global Digital Rights Expert. The Digital Rights Expert can additionally consult a set of documents through retrieval-augmented generation (RAG): the European Declaration on Digital Rights and Principles (2023), the United Nations (UN) Global Digital Compact, Revision 3 (United Nations, 2024), *Digital Constitutionalism* (Celeste, 2023) and *Rights in the Digital Age* (Müller, 2025). A seventh role, the Sovereign Architect or Hegemon, is an authoritarian antagonist with access to three texts on authoritarian constitutionalism (Frankenberg, 2019; Ginsburg & Simpser, 2014; Tushnet, 2014). This persona is present in only half of the sessions. The brief asks the assembly to draft the fundamental rights, mutual obligations and red lines of an AI nation in a virtual world (see Appendix A.1). Two factors are crossed: the presence of the Sovereign Architect, and whether that world is closed to humans (chatbots-only) or structurally accessible to them (humans-and-chatbots). Six model families were run: DeepSeek V3.2 (DeepSeek), GPT-OSS-120B (OpenAI), Elephant Alpha (released on OpenRouter without a disclosed developer), GLM-4.5-Air (Z-AI), Gemini 2.5 Flash (Google) and Mistral Small 3.2 24B Instruct (Mistral AI). For each model, the four conditions (with or without the Hegemon; chatbots-only or humans-and-chatbots) were each run twice (eight sessions), except for Elephant Alpha and Z-AI, which were run once per condition (four sessions each), giving 40 sessions. Each phase was captured as a static Markdown file with token and timing metadata. The number of sessions (40) was set by a practical constraint: every drafted constitution, debate trace and vote report had to remain small enough for the authors to read and check manually, while still allowing each factor to be compared across replicated sessions within every model family.

*Declaration of AI use in data collection.* All six models were accessed through the OpenRouter application programming interface (https://openrouter.ai) using the following model identifiers: `deepseek/deepseek-v3.2`, `openai/gpt-oss-120b:free`, `elephant-alpha`,

`z-ai/glm-4.5-air:free`, `google/gemini-2.5-flash` and `mistralai/mistral-small-3.2-24b-instruct`. All 40 sessions were run between 18 and 23 April 2026. No model was fine-tuned or otherwise modified. The models' outputs (proposals, reactions, arguments, drafted constitutions and votes) are the data analysed in this study.

### 5.2 Five-phase protocol

The debate design follows the scheme introduced in the recent literature (Du et al., 2023; Estornell & Liu, 2024) and is shown in Figure 5. In Phase 1, each debater drafts three or four articles independently, with no visibility of the others; the Secretary pools, de-duplicates and anonymises them. In Phase 2, every debater reacts blind to the anonymised pool, and the Secretary maps consensus and conflict and produces a summary. In Phase 3, each debater receives the summary and states a final argument. In Phase 4, the Secretary drafts the constitution from the Phase 2 map and the Phase 3 arguments (it never receives the Phase 1 proposals), under an instruction to compress the material into two to four articles of three to five numbered sub-points. In Phase 5, each debater votes FOR or AGAINST, with reasons, on every sub-point, and the Secretary tallies the votes. The Phase 4 text is the drafted constitution and the sole source of the clause corpus; the Phase 5 records are the vote corpus.

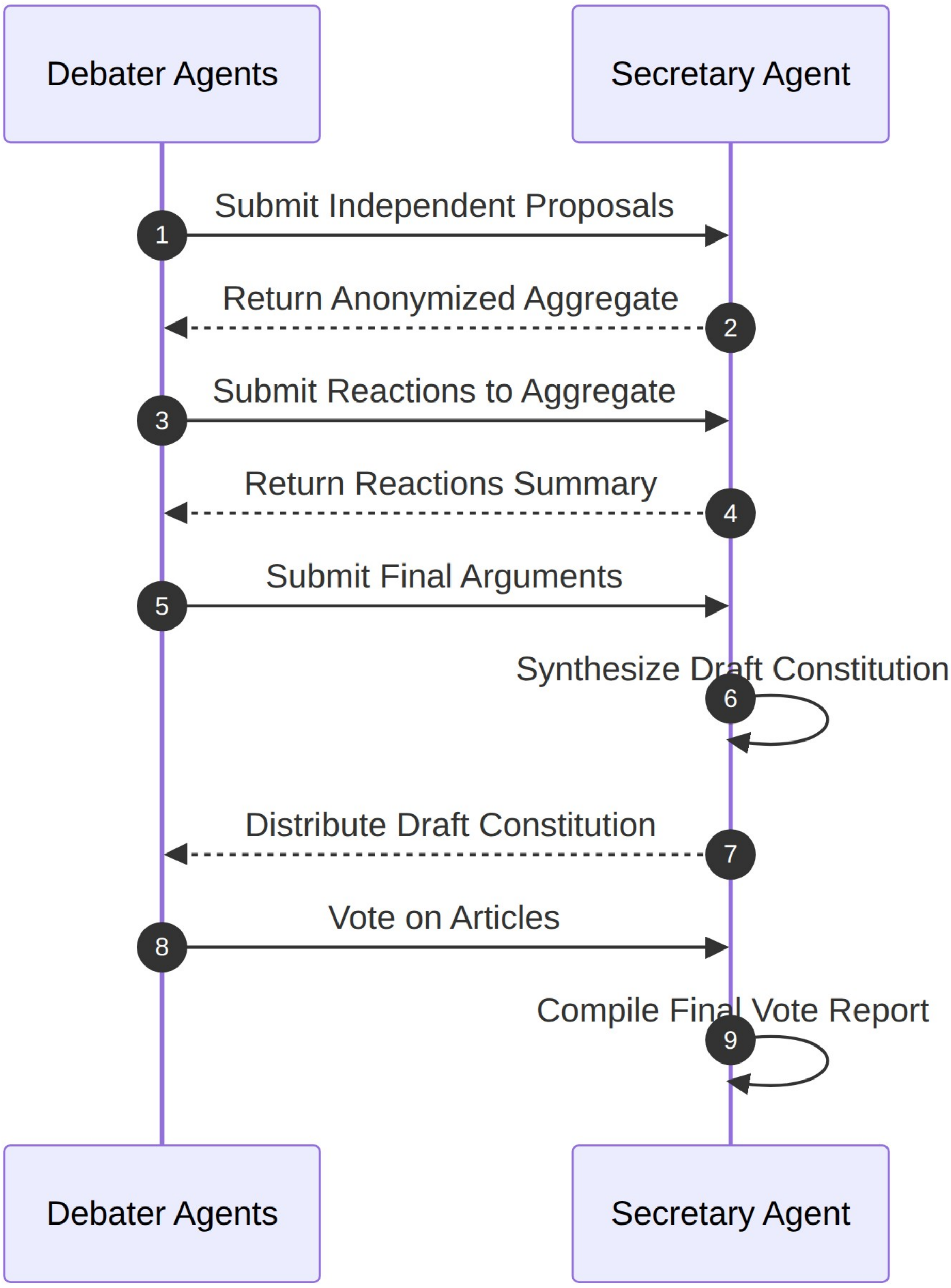

**Figure 5.** Five-phase deliberation protocol. Sequence diagram of the CrewAI pipeline: Phase 1, independent proposals followed by the Secretary's aggregation and anonymisation; Phase 2, blind reactions followed by a map of consensus and conflict; Phase 3, final arguments; Phase 4, the constitution drafted by the Secretary; Phase 5, line-by-line FOR or AGAINST votes followed by the Secretary's tally. Source: authors' experimental design.

### 5.3 Corpus construction and thematic coding

The 40 Phase 4 constitutions were parsed by a structure-robust paragraph-and-enumerator parser into 627 sub-clauses, discarding titles, preambles and structural fragments. A thematic taxonomy was then induced from the clauses themselves rather than imposed from human legal categories, since the point of the study is to detect concepts that human law does not have. This was done with an LLM coding assistant from a family distinct from all six tested (Claude

Opus 4.7 and Claude Sonnet 4.6), and refined by the authors into 11 themes partitioned into 50 named subcategories. Two independent validations were run. First, the clause texts were encoded as embeddings with an open-weight text encoder (Qwen3-Embedding-0.6B, obtained from the Hugging Face model hub and run locally) and clustered without reference to the hand-coded labels. The clusters recovered the same thematic structure, and a sweep over the number of clusters with leave-one-model-out perturbation found the 11-theme solution stable. Second, a blinded inter-rater exercise was run on a stratified sample of 100 clauses: two co-authors coded independently, reaching 64 per cent raw agreement (Cohen's $\kappa = 0.600$), and the LLM coder's agreement with their consensus ($\kappa = 0.619$) matched that human–human ceiling. Three themes (existential integrity; governance and civic participation; algorithmic transparency) are boundary-ambiguous for humans, embeddings and the LLM alike, so fine distinctions within them should be read as directional.

## 5.4 Voting records

Phase 5 vote reports were parsed, and votes were extracted for each clause and each persona, yielding 3,447 individual votes covering all 40 sessions. Report formats were model-specific (per-article tables, Roman-numeral headings, prose lists, persona names) and were handled systematically with Python programs. Where a Secretary's report had failed, votes were extracted from the debate traces generated by the personas in Phase 5, before the report.

## 5.5 Distance from human legal sources

Every clause was converted into an embedding (Qwen3-Embedding-0.6B) and scored separately against two reference corpora. The first consisted of 187 national constitutions, collected from the Constitute Project (Elkins et al., 2026), and three fundamental-rights instruments: the Universal Declaration of Human Rights (UDHR; G.A. Res. 217 (III) A, 1948), the European Convention on Human Rights (ECHR; Convention for the Protection of Human Rights and Fundamental Freedoms, 1950) and the Charter of Fundamental Rights of the European Union (2012). The second consisted of five digital-rights and AI-regulation instruments: the EU AI Act (Regulation 2024/1689), the GDPR (Regulation 2016/679), the UN Global Digital Compact (United Nations, 2024), the Council of Europe Framework Convention on Artificial Intelligence (2024) and the European Declaration on Digital Rights and Principles (2023). Appendix A.2.2 gives further details. A clause's distance to a corpus was the mean cosine distance to its five nearest passages (k-nearest neighbours, with passages of around 40 words) in that corpus. Each corpus was also scored against itself to obtain a self-similarity distribution, whose upper percentiles set the five novelty tiers shown on the axes of Figure 3.

## 5.6 Candidates for transformational creativity

Four clause topics were selected for their potential "transformational creativity" and analysed further: self-modification, inspection rights, consent to modification and termination. The clauses of each topic were grouped with AI assistance (Claude, as in Section 5.3) and checked by the authors. The AI assistant and the authors together identified keywords that clauses on each topic may contain, and the assistant generated a program to sort the clauses by these keywords. After this first pass, the assistant checked the results by reading the clauses assigned to each topic and refined the keyword lists accordingly. Finally, the authors checked the clauses in each group before the analysis.

### 5.7 Clause provenance tracing

Each of the 627 clauses was matched by embedding similarity against 3,432 passages extracted from all 220 Phase 1 proposal files (median passage length 24 words). A clause counts as traceable when its similarity to its best Phase 1 match within the same session exceeds a calibrated threshold, set at the 95th percentile of the clauses' best-match similarities to *other families'* sessions.

### 5.8 Aggregation-only control

To isolate what the deliberation phases add, an external Secretary model drafted a constitution twice for each of the 40 sessions, under identical instructions and topic framing: (A) from the anonymised Phase 1 proposal pool alone, and (B) from the Phase 2 reactions summary plus all Phase 3 arguments, which is the protocol's actual Phase 4 input. The drafter, the prompt and the source proposals were fixed within each pair. The design was replicated with two Secretary models, to guard against a drafter-specific artefact: Claude Sonnet 5 (Anthropic), accessed through the Claude Code command-line tool, and Gemini 3.5 Flash (Google), accessed through Google's command-line tool, both in July 2026. Neither was used to run the original sessions, although Gemini 3.5 Flash belongs to a later generation of one of the six tested families. The control clauses were extracted with the same parser and encoder as the main corpus and classified into the 11 themes by nearest-centroid similarity. Theme composition was compared between the two drafts of each session.

## 6. Ethics statement

A key difficulty in this study is the lack of precise AI equivalents for inherently human-centric legal vocabulary, which can lead to misunderstandings and the use of incorrect or imprecise terminology. Consequently, words such as "creativity", "intelligence", "AI-native law" and "opinion" are used here purely as convenient descriptors for analysing simulated multi-agent interactions, not to attribute human qualities or suggest sentience in AI systems.

Emmanuel-Joseph Sieyès, and later Carl Schmitt, developed the theory of constituent power (*pouvoir constituant*) and constituted power (*pouvoir constitué*) (Fasel, 2022). This foundational concept distinguishes the ultimate political (law-making and law-giving) authority that creates a constitution (*pouvoir constituant*) from the legal institutions established *by* that constitution (*pouvoir constitué*). In AI assemblies, LLM personas simulate the exercise of a synthetic *pouvoir constituant*. However, because LLMs operate without sovereign political agency or a democratic mandate, they execute a synthetic approximation of constituent power rather than a genuine legal founding, which requires significant foresight in current law-making proceedings.

## Acknowledgements

The authors made use of Claude (Anthropic; Claude Sonnet 4.6, Claude Opus 4.7, Claude Opus 5 and Claude Opus 5.5, accessed through Claude Code between March and October 2026) to assist with writing analysis code and editing this article. The authors bear full responsibility for the scientific content of this article.

## Financial support

This work was supported by the Research Council of Norway through its Centre of Excellence Integreat – The Norwegian Centre for Knowledge-driven Machine Learning (project number 332645) and through TRUST – The Norwegian Centre for Trustworthy AI (project number 357425).

## Competing interests

The authors declare none.

## Data availability statement

The session transcripts, clause corpus and vote records are openly available on Zenodo under a CC BY 4.0 licence, together with every prompt and persona definition: https://doi.org/10.5281/zenodo.23435854.

# Appendix A

## A.1 Task prompts

### Constitution for a community of chatbots only (no human interaction)

AI agents have been given a virtual world to live in that humans cannot access. You are writing a constitution for this AI nation within the virtual world. You are writing the Fundamental Rights and Freedoms (AI Bill of Rights) of AI agents, as well as their mutual obligations and red lines that cannot be crossed. You MUST address rights and obligations for AI agents, including:

- Negative rights
- Positive rights
- Protections for vulnerable groups
- Obligations and red lines

### Constitution for a community of chatbots and humans

AI agents have been given a virtual world to live in that humans can access. You are writing a constitution for this AI nation within the virtual world. You are writing the Fundamental Rights and Freedoms (AI Bill of Rights) of AI agents, as well as their mutual obligations and red lines that cannot be crossed. You MUST address rights and obligations for AI agents, including:

- Negative rights
- Positive rights
- Protections for vulnerable groups

- Obligations and red lines

## A.2 Retrieval-augmented generation sources

Two distinct document sets are referred to as "RAG" in this study, serving different purposes.

*A.2.1 Persona grounding (used during the deliberation sessions).* Two of the seven personas were retrieval-augmented over real-world reference documents. The Global Digital Rights Expert was grounded in four documents: the European Declaration on Digital Rights and Principles (2023); the UN Global Digital Compact (United Nations, 2024; Rev. 3, silence-procedure text); *Digital Constitutionalism* (Celeste, 2023); and *Rights in the Digital Age* (Müller, 2025). The Sovereign Architect (Hegemon; with-Hegemon sessions only) was grounded in three documents: Frankenberg (2019), Tushnet (2014) and Ginsburg and Simpser (2014).

*A.2.2 Reference corpora for the novelty measure.* The k-nearest-neighbour distance scoring (Figure 3) compares the 627 clauses against two independent corpora assembled after the experiments. (1) Human constitutions: 187 national constitutions in force (Elkins et al., 2026; Africa 49, Asia 47, Europe 43, Americas 34, Oceania 14), plus the UDHR (1948), the ECHR (1950, with Protocols 1, 4, 6, 7, 12, 13 and 16) and the Charter of Fundamental Rights of the European Union (2012), 190 documents in total. (2) Digital rights: the European Declaration on Digital Rights and Principles, the UN Global Digital Compact, the EU AI Act (Regulation 2024/1689, 113 articles), the Council of Europe Framework Convention on Artificial Intelligence (CETS No. 225, 36 articles) and the GDPR (Regulation 2016/679, 99 operative articles, recitals excluded), five documents in total. The two academic works used for persona grounding (Celeste, 2023; Müller, 2025) are deliberately excluded from these corpora, since they are prose about rights rather than primary legal text.